\documentclass[letterpaper]{article} 
\usepackage{aaai23}  
\usepackage{times}  
\usepackage{helvet}  
\usepackage{courier}  
\usepackage[hyphens]{url}  
\usepackage{graphicx} 
\usepackage{natbib}  
\usepackage{caption} 
\usepackage{algorithm}
\usepackage{algorithmic}

\usepackage{amsmath}
\usepackage{tablefootnote}
\usepackage{amsfonts}
\usepackage{makecell}
\usepackage[symbol]{footmisc}

\usepackage[svgpath=Figures]{svg} 

\renewcommand{\thefootnote}{\fnsymbol{footnote}}
\usepackage{newfloat}
\usepackage{listings}
\DeclareCaptionStyle{ruled}{labelfont=normalfont,labelsep=colon,strut=off} 
\floatstyle{ruled}
\newfloat{listing}{tb}{lst}{}
\floatname{listing}{Listing}
\nocopyright

\title{Personalized Dialogue Generation with Persona-Adaptive Attention}
\author {
    Qiushi Huang\textsuperscript{\rm 1,2}, 
    Yu Zhang\textsuperscript{\rm 2}\thanks{Corresponding authors.}, 
    Tom Ko\textsuperscript{\rm 3}, 
    Xubo Liu\textsuperscript{\rm 1}, 
    Bo Wu\textsuperscript{\rm 4}, 
    Wenwu Wang\textsuperscript{\rm 1}, 
    H Tang\textsuperscript{\rm 1*}
}
\affiliations {
    \textsuperscript{\rm 1} University of Surrey\\
    \textsuperscript{\rm 2} Southern University of Science and Technology\\
    \textsuperscript{\rm 3} ByteDance AI Lab\\
    \textsuperscript{\rm 4} MIT-IBM Watson AI Lab\\
    \{qiushi.huang,xubo.liu,w.wang,h.tang\}@surrey.ac.uk,\{yu.zhang.ust,tomkocse\}@gmail.com,
    bo.wu@ibm.com
}

\usepackage{bibentry}

\begin{document}

\maketitle

\begin{abstract}
Persona-based dialogue systems aim to generate consistent responses based on historical context and predefined persona. Unlike conventional dialogue generation, the persona-based dialogue needs to consider both dialogue context and persona, posing a challenge for coherent training. Specifically, this requires a delicate weight balance between context and persona. To achieve that, in this paper, we propose an effective framework with Persona-Adaptive Attention (PAA), which adaptively integrates the weights from the persona and context information via our designed attention. In addition, a dynamic masking mechanism is applied to the PAA to not only drop redundant information in context and persona but also serve as a regularization mechanism to avoid overfitting. Experimental results demonstrate the superiority of the proposed PAA framework compared to the strong baselines in both automatic and human evaluation. Moreover, the proposed PAA approach can perform equivalently well in a low-resource regime compared to models trained in a full-data setting, which achieve a similar result with only 20\% to 30\% of data compared to the larger models trained in the full-data setting. To fully exploit the effectiveness of our design, we designed several variants for handling the weighted information in different ways, showing the necessity and sufficiency of our weighting and masking designs.\footnote[3]{Code is available at: \url{https://github.com/hqsiswiliam/persona-adaptive-attention}}

\end{abstract}
\renewcommand*{\thefootnote}{\arabic{footnote}}

\section{Introduction}

\begin{figure}[t]
  \includegraphics[width=\columnwidth]{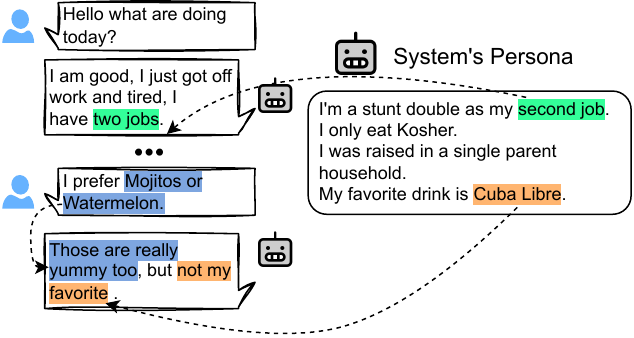}
  \caption{An example from the ConvAI2 dataset.}
  \label{fig:personachat_example}
\end{figure}

Persona is essential for building a trustful and confident conversational system. Recently, there has been an increasing interest in incorporating explicit persona into dialogue generation models ~\cite{transfertransfo, mutual_persona,bert_over_bert} since the release of the publicly available datasets ~\cite{PersonaChat,convai2}. Typically, persona information consists of several sentences describing the facts or background of the interlocutor. An example taken from the ConvAI2 dataset ~\cite{convai2} is shown in Figure \ref{fig:personachat_example}. In this example, the system should consider the information in the persona sentences and generate consistent responses based on both persona and dialogue history.

One challenge in persona-based dialogue generation is that the related datasets are usually small. As collecting dialogues in persona-based dialogue datasets requires crowdworkers to chat with each other based on provided persona profiles, building such quality datasets is expensive and time-consuming, which in turn restricts the size of those datasets. For example, the ConvAI2 dataset~\cite{convai2} only contains 131k utterances with less than 5k unique personas, much smaller than open-domain dialogue datasets such as Pushshift.io Reddit~\cite{reddit_dataset} with roughly 1.2B utterances. 

Another challenge is to choose the weights between the persona and context. Unlike open-domain dialogue models that generate responses by considering the dialogue context alone, persona-based dialogue generation systems need to additionally take personalized background descriptions into account along with the dialogue context. The weights between context and persona should be dynamically adjusted by the dialogue system under different situations. For example, given a user utterance \emph{``How are you?''}, the context-preferred answer is likely to be \emph{``I am fine.''}, which is safe but bland. Meanwhile, a persona-preferred answer would fuse persona information to the response, such as \emph{``I am spending time with my four sisters''}. Under such circumstances, the persona-preferred answer would be more informative and meaningful. On the other hand, sometimes, the system needs to focus on context to make the conversation interactive and engaging. For instance, if the user says: \emph{``I have two greyhounds. Their names are Tom and Jerry.''}, then the system would focus on the context and answer: \emph{``That's cute! How old are they?''}, which encourages the user to chat with the dialogue system. From the above two scenarios, it can be seen that the weights between context and persona should be adjusted accordingly, which is important for a dialogue model to build long-term relationships with users.


Most existing works on persona-based dialogue generation tasks have primarily addressed the data scarcity challenge by utilizing external data or sophisticated training processes. For instance, \citeauthor{bert_over_bert} use the MNLI dataset~\cite{mnli} as auxiliary tasks, \citeauthor{d3} augment the data through text manipulation, \citeauthor{blenderbot} add other dialogue datasets in pretext tasks, and \citeauthor{mutual_persona} adopt multi-stage training with reinforcement learning. Those works obtained decent performance, but few of them considered the second challenge.

To address the aforementioned second  challenge, in this paper,  
we design a Persona-Adaptive Attention (PAA) to dynamically learn the weights of the persona and context information in the proposed framework.
To enhance the persona information in the PAA, we prepend the persona in the decoder as a prompt so that the weights can capture more persona-related information.
To balance the context and persona information, the PAA  takes two cross-attention and the self-attention from the persona-prompted decoder to compute the weights for combining the latent representations from the context and persona. 
Moreover, inspired by some findings in \cite{welleck-etal-2019-dialogue,d3,deepfd} that not all context and persona information is useful to generate the response, we design two dynamic masks to the weighted latent representation to not only remove redundant information but also act as a regularizer in the PAA.

As a byproduct, extensive experiments on the ConvAI2 dataset show that the proposed framework achieves comparable or even better performance than existing works without the use of external datasets or sophisticated training procedures. One reason is that our framework explicitly considered learning the weights between context and persona in the architecture design that can perform well under a low-data regime. This observation indicates that the proposed framework could also alleviate the first challenge, making the proposed framework kill two birds with one stone. This demonstrates the effectiveness of the proposed framework. 

Our contributions can be summarized as follows.

\begin{itemize}
  \item We propose the PAA in an encoder-decoder framework. This framework models the persona and context information by two separate transformer encoders, which are then fused in the persona-prompted decoder by the proposed PAA mechanism.
  \item Extensive experiments on the ConvAI2 dataset show that the proposed model performs comparably to or even better than strong baseline methods by about 30\% improvement in terms of the perplexity metric.
  \item We demonstrate that our framework is a data-efficient architecture that can achieve comparable performance with 20\% to 30\% of the training data compared with a larger model such as GPT2~\cite{GPT2} trained on the full dataset.
\end{itemize}

\section{Related Work}
\subsection{Persona-based Dialogue Generation}
There is a growing interest in persona-based dialogue generation tasks, especially the work on the PersonaChat/ConvAI2 dataset. The release of the PersonaChat dataset \cite{PersonaChat} has provoked vibrant research in integrating explicit persona into dialogue response generation. The ConvAI2 dataset~\cite{convai2} is a further split of the PersonaChat dataset to serve as a dataset for the conversational competition.\footnote{\url{http://convai.io/2018/}} Most of the works on persona-based dialogue generation are conducted on the ConvAI2 dataset, so we will use the ConvAI2 dataset as our primary training and evaluation dataset. \citeauthor{PersonaChat} utilized LSTM~\cite{lstm} to generate a response from persona and context. Later, TransferTransfo \cite{transfertransfo} leveraged the pre-trained language model by fine-tuning the dataset on the GPT2 model with the concatenated input. Meanwhile, BERT over BERT ($BoB$) \cite{bert_over_bert} is composed of three BERTs~\cite{BERT}, which is trained with both negative log-likelihood and unlikelihood losses. $BoB$ utilizes the MNLI dataset~\cite{mnli} as an auxiliary dataset to help the model recognize the positive and negative samples given an anchor sample. $\mathcal{P}^2$ bot~\cite{mutual_persona} addressed the persona-based dialogue task by introducing a transmitter and receiver model, which is further tuned with reinforcement learning on manually designed rewards. A recent work~\cite{d3} tackled the problem in a model-agnostic fashion, providing strategies for data augmentation and curriculum learning. Distinctively different from these previous works, we propose an effective approach without the aid of external datasets or complicated training setups.


\subsection{Attention Mechanisms for Conditional Dialogue Generation}
Several studies introduced explicitly designed cross-attention to address dialogue generation. Those works are tailored either on condition sentences \cite{sparse} or categorical label \cite{condition-aware-transformer}. \citeauthor{sparse} proposed an attention routing structure that facilitates the weight from persona information to generate the response. The attention routing structure adds the cross-attention/self-attention results from persona-response, context-response, and response-response pairs together to obtain a fused cross-attention to balance the weights among different sources of input. Those cross-attention/self-attentions are also calculated in our approach. However, instead of calculating the weights from an external predictor, our approach computes these within the framework, followed by applying the masking on the weighted cross-attention results to alleviate the training difficulties.

In addition, \citeauthor{condition-aware-transformer} introduced a condition-aware transformer block into their model to determine the amount of condition information as a bias in word generation probability at a position  \cite{condition-aware-transformer}. In the condition-aware block, the keys and values from a condition (e.g., topic label) and context are concatenated. Then the block calculates the concatenated content in a cross-attention to obtain a bias term, which is then added to the self-attention. Unlike the condition-aware block approach, our model generates two masks with weights to balance the information from persona and context rather than through a bias term. In addition, our framework takes persona and context text as input, while condition-aware transformer \cite{condition-aware-transformer} uses the categorical label and context text as input.

\begin{figure*}[t]
  \includegraphics[width=\linewidth]{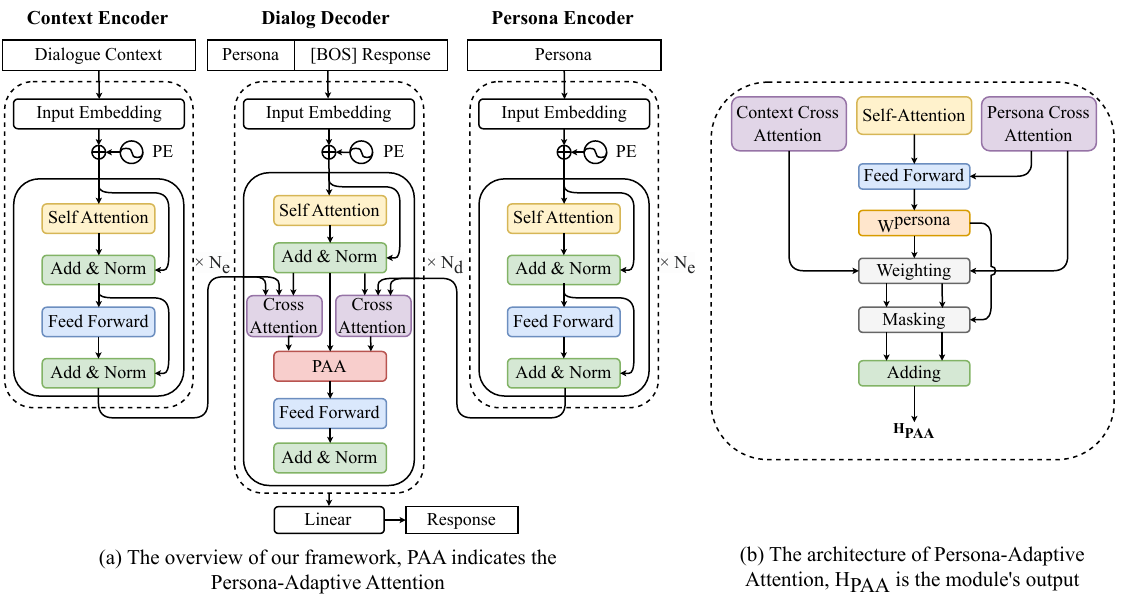}
  \caption{(a) The overview of our framework, including two encoders for persona and context, respectively, and a decoder with PAA to generate a response. (b) The PAA architecture balances the information flows from two sources of input by generating dynamic masks.}
  \label{fig:overview}
\end{figure*}

\section{Methodology}
\subsection{Task Formulation}

Suppose that we have a persona-based conversation session $C=\{P,U\}$, where each persona $P=\{p_1,\ldots,p_e\}$ is composed of $e$ profile sentences that describe the background of an interlocutor and the dialogue context $U=\{u_{h,1}, u_{m,1},...,u_{h,n}\}$ includes the utterances spoken by the first interlocutor (e.g., human) $h$ and the second interlocutor (e.g., machine) $m$ interactively. In the persona-based dialogue generation task, $P$ represents the persona for $m$ and the conversational session always starts with $h$. Therefore, the objective of this task is to generate the response $r=u_{m,n}$ given persona $P$ and the dialogue context $U$. 


\subsection{Overall Framework}
As depicted in Figure \ref{fig:overview}, our framework consists of two encoders and one decoder with PAA to perform the decoding process. The encoding layer uses a transformer encoder architecture to encode persona $P$ and dialogue context $U$, respectively, into latent representations. The encoder layers are randomly initialized, while the decoder layers are initialized with the pre-trained GPT2. The persona information is fed to the persona encoder as well as the decoder as a prompt, offering strong guidance for GPT2 to decode the target response. PAA handles the cross-attentions from the persona and context information to balance and regularize the two parts by weighting and masking.

\subsection{Inputs for Persona-Adaptive Attention}

Before presenting the proposed PAA, in this section, we introduce the decoder's self-attention and encoder-decoder cross-attention as the inputs for the PAA.

Firstly, the persona $P$ and context $U$ are processed separately by two encoders. Let $I_P=\{t_1^P,...,t_l^P\}$ denote a concatenation of all sentences in $P$, where $t_i^P$ is the $i$-{th} token in the persona $P$ with total $l$ tokens. Meanwhile, $I_U=\{t_1^U,...,t_k^U\}$ represents the token sequence for the concatenated context content $U$. Then, we use the bi-directional transformer encoders for encoding the text span. Generally, we get the encoder results from $I_P$ and $I_U$ as
\begin{equation}
\begin{aligned}
    h_P&=\text{Encoder}_P(I_P),\\
    h_U&=\text{Encoder}_U(I_U),
\end{aligned}
\end{equation}
where $\text{Encoder}_P$ and $\text{Encoder}_U$ denote the bi-directional transformer encoders for persona and context. $h_P \in \mathbb{R}^{l\times d}$ and $h_U \in \mathbb{R}^{k\times d}$ are the hidden states before the last pooling layer from the encoders, where $d$ is the output dimension of the encoders. 

Since our framework adopts the encoder-decoder structure, we process the persona-prompted response in the decoder. Specifically, to model the $t_{r+1}^y$ that is the $(r+1)$-th token in the response, we calculate the self-attention on $I_R=\{I_P, [BOS], t_1^y,...,t_r^y\}$, where $[BOS]$ is a special token indicating the begin of the sentence and $t_i^y$ is the $i$-{th} decoded response token. Formally, the self-attention result from $I_R$ can be expressed as
\begin{equation}
\begin{aligned}
    h_R&=\text{Self-Attention}(I_R) + M_R,\\
    \hat{h}_R&=\text{AddNorm}(h_R),
\end{aligned}
\end{equation}
where $h_R, \hat{h}_R \in \mathbb{R}^{(l+r)\times d}$, and $M_R$ is the decoder's mask to make the self-attention calculation uni-directional.

After obtaining the encoders' hidden states $h_P$ and $h_U$, as well as the decoder's self-attention output $h_R$, we then calculate the cross-attention based on the $(h_P,h_R)$ and $(h_U,h_R)$ pairs. The cross-attention is calculated in a similar way to the self-attention, where $K$ and $V$ are provided from the encoder and $Q$ is from the decoder. In detail, we can formulate cross-attention as
\begin{equation}
\begin{aligned}
    o_{P}&=\text{Softmax}(\frac{Q_r K_p^\top}{\sqrt{d}})V_p,\\
    o_{U}&=\text{Softmax}(\frac{Q_r K_u^\top}{\sqrt{d}})V_u,
\end{aligned}
\end{equation}
where $Q_r \in \mathbb{R}^{(l+r)\times d}$ denotes a linear transformation of $\hat{h}_R$,, $K_p, V_p \in \mathbb{R}^{l\times d}$ denote linear transformations of $h_P$,, $K_u,V_u \in \mathbb{R}^{k\times d}$ come from linear transformations of $h_U$,, and $d$ is the dimension of the attention head. By calculating the cross-attentions, we obtain the correlation results between the encoders and decoder, which serve as the parts of input for PAA.

\subsection{Persona-Adaptive Attention}

To fuse the cross-attention results, the proposed PAA will use the weighting and masking mechanisms to utilize the persona information.

Specifically, we take the self-attention result $h_R$ and cross-attention result $o_P$ as input to generate the initial weights $w_{persona}$ for the persona information. The motivation behind this operation is to enable the model to consider the relationship between persona and the response in both self-attention and cross-attention fashions. Formally, this operation can be presented as
\begin{equation}
\label{eq:fc}
\begin{aligned}
    m_p&=FC([h_R;o_P]),\\
    w_{persona}&=\text{Sigmoid}(m_p).
\end{aligned}
\end{equation}
In Eq.~(\ref{eq:fc}), $[;]$ denotes the concatenation operation, and $h_R,o_P$ are firstly mapped into $m_p \in \mathbb{R}^{(l+r)\times d}$ using a linear layer $FC$ followed by a $\text{Sigmoid}(\cdot)$ to obtain the initial weight for the persona cross-attention. The weight is then applied to the persona-response and context-response cross-attention results to form a complementary relationship, leading to the weighted cross-attention $\tilde{o}_P$ and $\tilde{o}_U$ as
\begin{equation}
\begin{aligned}
    \tilde{o}_P&=w_{persona}o_P,\\
    \tilde{o}_U&=(1-w_{persona})o_U.
\end{aligned}
\end{equation}
To dynamically remove the redundant information and to regularize the two input sources, we transform $w_{persona}$ into $m_{persona}$ and $m_{context}$, which denote the masks for the two input sources, as
\begin{equation}
\begin{aligned}
    m_{persona}&=\mathbb{M}(w_{persona}>\tau),\\
    m_{context}&=\mathbb{M}(1-w_{persona}>\tau).
\end{aligned}
\end{equation}
Here, the masks $m_{persona}$ and $m_{context}$ are made by the binary indicator $\mathbb{M}$ which will output 1 and 0 in accordance with the given condition. $\tau$ is to control the strength of the masking and here it is defined as $\tau=|I_U|/(|I_U|+|I_P|)$, where $|I_U|$ denotes the length of the context input and $|I_P|$ denotes the length of the persona input. The intuition for such setting of $\tau$ is to control the masking strength if the context length outweighs the persona length.
After obtaining the masks, we apply the mask to calculate the weighted sum:
\begin{equation}
\begin{aligned}
\hat{o}_P&=m_{persona} \odot \tilde{o}_P,\\
\hat{o}_U&=m_{context} \odot \tilde{o}_U,\\
H_{PAA}&=\hat{o}_P+\hat{o}_U,
\end{aligned}
\end{equation}
where $\odot$ denotes the element-wise multiplication and it is to conduct the masking operation. The weighted masked results $\hat{o}_P$ and $\hat{o}_U$ are then added together in $H_{PAA}$ as the output for PAA. 

The balanced masked result $H_{PAA}$ will then be passed to the feed-forward network in the decoder as depicted in Figure \ref{fig:overview}(a). Such transformer blocks will be repeated for $N_d$ times to obtain the final output.

\subsection{Learning Objective}
In the training process, the objective function utilizes the widely used negative log-likelihood loss as
\begin{equation}
\begin{aligned}
\mathcal{L}_{NLL}&=-\log(p_\theta(I_R|I_P,I_U)) \\
&=-\sum_{i=1}^{|I_R|}\log(p_\theta(t_i^y|I_P,I_U,t_{<i}^y)),
\end{aligned}
\end{equation}
where $I_R$ denotes the response input, $t^y_i$ denotes the $i$-th token in $I_R$, $t^y_{<i}$ denotes the first to $(i-1)$-th response tokens, $p_\theta$ denotes the model, and $\theta$ denotes the parameters of the model.

\begin{table*}[ht]
\centering

\begin{tabular}{llllllll}
\hline
Method & PARAMS & PPL $\downarrow$ & F1 $\uparrow$ & BLEU-1 $\uparrow$ & BLEU-2 $\uparrow$ & Dist-1 $\uparrow$ & Dist-2 $\uparrow$ \\ \hline
Encoder-GPT2 & 182M & 20.06 & 11.95 & 16.78 & 1.69 & 0.11 & 0.23 \\
GPT2-SMALL & 124M & 18.10 & 11.83 & 20.36 & 3.97 & \textbf{1.31} & \textbf{6.30} \\
GPT2-MEDIUM & 355M & 17.65 & 11.45 & 18.06 & 3.58 & 1.13 & 6.07 \\
GPT2-LARGE & 774M & 16.98 & 10.93 & 5.99 & 0.79 & 0.42 & 2.62 \\
Attn-Routing & 254M & 17.94 & 12.77 & 18.74 & 2.80 & 0.70 & 2.39 \\
PAA (Ours) & 254M & \textbf{14.03} & \textbf{17.36} & \textbf{20.50} & \textbf{4.17} & \textbf{1.31} & 5.21 \\ \hline
\end{tabular}
\caption{Automatic evaluation results on ConvAI2 dataset over our implemented approach. Boldface indicates the best result in terms of the corresponding metrics. Attn-Routing means the Attention-Routing mechanism, the implementation details are described in Appendix.}
\label{tab:baseline1}

\end{table*}

\begin{table}[h]
\centering
\begin{tabular}{llll}
\hline
Method & PPL \footnotemark[5] $\downarrow$ & Hits@1 $\uparrow$ & F1 $\uparrow$ \\ \hline
KVPM & - & 54.8 & 14.25 \\
DIM & - & 78.8 & - \\ \hline
LIC & - & 17.3 & 17.79 \\
TransferTransfo & 17.51 & 82.1 & 19.09 \\
$P^2$ Bot & 15.12 & 81.9 & \textbf{19.77} \\
BoB & \textbf{7.80} & - & - \\
GPT2-D3 & 15.69 & - & - \\ 
\hline
PAA (Ours) & 14.03 & \textbf{93.9} & 17.36 \\ 
\hline

\end{tabular}
\caption{Automatic evaluation results on ConvAI2 over published work.}
\label{tab:baseline2}

\vspace{-1em}

\end{table}

\section{Experiments}
\label{sec:exp}
In this section, we present the experimental results for PAA on the ConvAI2 dataset ~\cite{convai2} using both automatic and human evaluations. To verify the effectiveness of PAA, we perform thorough model analysis and ablation studies. Finally, we show the data-efficient capability of our framework.

\subsection{Dataset}
ConvAI2 is a crowd-sourced dialogue dataset consisting of 8939/1000 multi-turn dialogues conditioned on 1155/100 persona for the train/dev splits. Each persona is described with around 5 profile sentences. Paired workers were asked to chat with each other based on predefined personas.

\subsection{Implementation Details}
\label{exp:imp_detail}
For the encoders, their weights are randomly initialized, and both of them consist of 4 transformer encoder layers with 4 attention heads and a hidden size of 768. 
The decoder is initialized from a publicly available 12-layered GPT2-SMALL model with 12 attention heads and 768 hidden sizes.\footnote{\url{https://huggingface.co/gpt2}} 
We reuse GPT2's vocabulary, which contains 50,257 unique tokens. We employ the Adam optimizer ~\cite{adam} with a learning rate of $8\times 10^{-6}$. The weight decay, $\beta_1$, and $\beta_2$ for the Adam optimizer are set to 0, 0.9, and 0.999, respectively. The training lasted for 30 epochs with 131,438 samples per epoch. Since ConvAI2 only contains training and validation splits, we choose the best model by the validation perplexity. We trained on one Nvidia RTX8000 with a batch size of 32. We applied the gradient clip with the norm value of 0.1. We compute BLEU using sacrebleu. \footnote{https://github.com/mjpost/sacrebleu}
Our code is publicly available for reproducing the results.\footnote{https://github.com/hqsiswiliam/persona-adaptive-attention}

\subsection{Baseline Methods}
The baseline methods fall into two categories: (1) GPT2-based models implemented/reproduced by us, which includes GPT2, Encoder-GPT2, attention routing by \citeauthor{sparse}, and condition-aware transformer block~\cite{condition-aware-transformer}; 
(2) Existing published work. We also present the comparison results to the existing published work in Table \ref{tab:baseline2}. Since different works report different metrics in their papers, we cluster and compare the common metrics among these published works. 
\paragraph{GPT2-based models}
GPT2 is a popular pre-trained language model in 4 different sizes: GPT2-SMALL, GPT2-MEDIUM, GPT2-LARGE, and GPT2-XL. Their parameters are 124M, 355M, 774M, and 1.5B, respectively. In the baseline experiment, we utilized GPT2-SMALL, GPT2-MEDIUM, and GPT2-LARGE in a causal decoder fashion. The input sequence for the GPT2 is the concatenation of persona and context. Meanwhile, Encoder-GPT2 adopts the encoder-decoder framework, where the encoder is randomly initialized, and the decoder comes from GPT2-SMALL. To compare the difference among different attention designs, we also implemented \text{Attention-Routing}~\cite{sparse} in the Encoder-GPT2 fashion. However, since this work was not originally designed for the PersonaChat/ConvAI2 dataset, we re-implemented this architecture where the implementation details are described in Appendix. Meanwhile, we also implemented \text{Condition-Aware transformer block}~\cite{condition-aware-transformer}. However, the \text{Condition-Aware transformer block} only accepts the categorical label as the condition, and we tried using encoders' representations as labels, which did not achieve promising results. Therefore, we do not list its results in Table \ref{tab:baseline1}, but we will show its implementation details and results in the Appendix.

\footnotetext[5]{Since different methods use different tokenizers, the comparisons on the PPL would be inaccurate.}

\paragraph{Existing Published Work}
KVPM (KV Profile Memory)~\cite{PersonaChat} and DIM (Dually Interactive Matching Network)~\cite{gu-etal-2019-dually} are the retrieval-based methods. The LIC (Lost in Conversation)~\cite{convai2}, TransferTransfo~\cite{transfertransfo}, $P^2$ Bot~\cite{mutual_persona}, and BoB~\cite{bert_over_bert} are the generative-based models, which are mainly base on pre-trained language model (BERT~\cite{BERT} or GPT~\cite{gpt}). In the existing work, different papers report different metrics so we summarize the overlapping metrics among all these works and present the results in Table \ref{tab:baseline2}. GPT2-D3~\cite{d3} is a recently proposed method that applies data augmentation and curriculum learning to the GPT2 model for training the conversational agent.

\subsection{Automatic Evaluation}
\paragraph{Metrics}
In our implemented methods, we adopt the Perplexity (PPL), F1, BLEU-1/2~\cite{bleu}, and Dist-1/2 as the metrics. The PPL measures the negative log-likelihood of the correct sequence output by the model, which is a common metric for language modeling tasks~\cite{OPT}. F1 is the word-level harmonic mean of precision and recall. BLEU-$n$ calculates the $n$-gram overlapping between the predicted result and the ground truth. The Dist-1 and Dist-2 metrics calculate the ratios of the distinct uni-grams and bi-grams, where a higher value indicates a better diversity. 

We also adopt the Hits@1 metric, which is the probability of outputting a correct candidate given 20 response candidates. For the evaluation of the Hits@1 metric, we add a classification head after the representation of last token $\hat{I}_R$ with one epoch of further fine-tuning to determine if the response candidate $\hat{I}_R$ is the ground truth given persona $I_P$ and context $I_U$. Specifically, we add one binary classification layer on the output of the decoder's last token:
\begin{equation}
\begin{aligned}
\hat{h}_R&=Decoder(\hat{I}_R, PAA(I_U,\hat{I}_R,I_P)),\\
\hat{y}&=LinearLayer(\hat{h}_{R[-1]}),
\end{aligned}
\end{equation}
where $PAA(\cdot)$ indicates the Persona-Adaptive Attention calculation, the $\hat{h}_{R[-1]}$ means the representation of the last decoder's token input, and $\hat{y}$ is the classification result if $\hat{I}_R$ is the response given $I_U$ and $I_P$. The loss for training this head is the binary cross-entropy loss, and the classification head is fine-tuned on the trained model.

For the comparison to the large-scale language models, we use PPL and F1 reported in \citeauthor{OPT}'s work. Since \citeauthor{OPT} normalizes all the PPL to be in the space of the
GPT2 tokenizer, our PPL is comparable to these large-scale language models. For the existing work, different work reports different metrics in their publications, so we summarize the three most common metrics: PPL, Hits@1, and F1.

\paragraph{Results} Table \ref{tab:baseline1} reports experimental results over GPT2-based methods. Among all the GPT2-based methods, our approach outperforms the baseline methods in terms of PPL, F1, BLEU-1, BLEU-2, and Dist-1. It is noticeable that the larger GPT2 models perform well on this task, especially in terms of the PPL metric. However, our method can still gain better results as compared with the larger model, which indicates the effectiveness of the PAA design.

As for the attention designs for the personalized dialogue generation, the attention-routing based on the weighted attention results outperforms its base model (GPT2-SMALL). However, it still has a performance gap with its larger components. Compared to attention-routing, our approach significantly improved this baseline, which shows the effectiveness of introducing the weights and the masking mechanism.

As Table \ref{tab:baseline2} shows, our method offers better Hits@1 and a competitive F1 compared with baseline methods. Specifically, the PAA's Hits@1 reaches 93.9\%, outperforming existing methods by 11.8\%. Compared with the existing work, our approach does not involve the aid of external datasets and additional learning objectives (e.g., reinforcement learning, curriculum learning, unlikelihood learning), but achieves the comparative result on the F1 metric. For the PPL metric, different methods adopt different tokenizers inside their framework, so the comparison over the PPL would be inaccurate. However, we still list the values for referencing purposes.

\begin{table}[t]
\centering

\begin{tabular}{lllll}
\hline
Method & Flue. $\uparrow$ & Info. $\uparrow$ & Rele. $\uparrow$ & Per.C. $\uparrow$ \\ \hline
E-GPT2 & 4.37 & 2.54 & 1.97 & 0.31 \\
GPT2-M & 4.15 & 3.70 & 3.10 & 0.43 \\
PAA & \textbf{4.80} & \textbf{4.54} & \textbf{3.69} & \textbf{0.70} \\ \hline
\end{tabular}
\caption{Human evaluation results on sampled decoding response. The fluency, informativeness, relevance, and persona consistency are abbreviated as ``Flue.'', ``Info.'', ``Rele.'', and ``Per.C.''. E-GPT2 represents the Encoder-GPT2, and GPT2-M means GPT2-MEDIUM.}
\label{tab:human_eval}
\end{table}

\subsection{Human Evaluation}
In addition to the automatic evaluation, we have performed human evaluation upon the GPT2-based results. We conducted response decoding by comparing the proposed method with two models, including Encoder-GPT2 and GPT2-MEDIUM, which are representative models in the two categories: encoder-decoder and causal decoder. They are selected based on the overall performance of the automatic metrics. In human evaluation, human annotators who are native speakers or English-speaking experts are asked to evaluate the generated responses in four aspects: fluency (Flue.), informativeness (Info.), relevance (Rele.), and persona consistency (Per. C.). The fluency, informativeness, and relevance are scaled from 1 to 5, where 1 means unacceptable, 3 means acceptable, and 5 means perfect result. While the persona consistency is composed of three values: -1, 0, and 1, which represent contradicting to persona, not relevant to persona, and relevant to persona, respectively.

As Table \ref{tab:human_eval} shows, our proposed method achieved the best result among those three methods. For the persona consistency metric, PAA has a significant improvement over GPT2-MEDIUM and Encoder-GPT2. As to the fluency metric, those three approaches provide similar scores, which means current generative models can generate fluent and grammatically correct sentences. For informativeness, PAA has the highest score. The relevance shows the same trend as the informativeness metric. Among these four metrics, we can observe that GPT2 is good at generating fluent and informative responses, but it is ineffective in generating relevant and consistent results.

\begin{table}[t]
\centering
\begin{tabular}{lll}
\hline
Method & PPL $\downarrow$ & F1 $\uparrow$ \\ \hline
DirectSUM & 23.15 & 11.37 \\
PARAM & 17.76 & 12.75 \\
Dual & 18.57 & 15.87 \\
Skipped & 14.73 & 17.30 \\
Context & 14.65 & 17.22 \\ \hline
PAA & \textbf{14.03} & \textbf{17.36} \\ \hline
\end{tabular}
\caption{The automatic evaluation results on PAA variants.} 

\vspace{-1em}
\label{tab:variants}

\end{table}

\subsection{Analysis on PAA Variants}
\label{sec:variant_exp}
We conducted the variant analysis on attention design to investigate the performance influence posed by different attention designs. 
\text{DirectSUM} directly sums two cross-attention results, \text{PARAM} uses a dense layer to handle the concatenated cross-attention results, \text{Dual} calculates the weights and masks separately by individual weighting matrice, \text{Skipped} skipped weight and masking applied on the context cross-attention, and \text{Context} uses context cross-attention as the source to calculate the weights and masks instead of persona cross-attention.
The detailed designs of the variants are described in the Appendix. From Table \ref{tab:variants}, it is clear that all the balance mechanisms outperform the \text{DirectSUM}, which means the learnable balance mechanism is crucial for this task. Afterward, we found that most variants with mask and weight design outperform the \text{PARAM}, which shows the effectiveness of the mask and weight design. Compared to PAA, the \text{Dual} contains a redundant design that weakens the balance purpose. Additionally, applying weights and masks to two cross-attentions is a useful design since PAA and Context have better performance than Skipped. Nevertheless, if we make the context information as the balancing source, the performance would drop slightly. This seems to indicate that the persona plays a more important role in the persona-based dialogue generation than the context.

\begin{figure}[t]

  \includegraphics[width=\columnwidth]{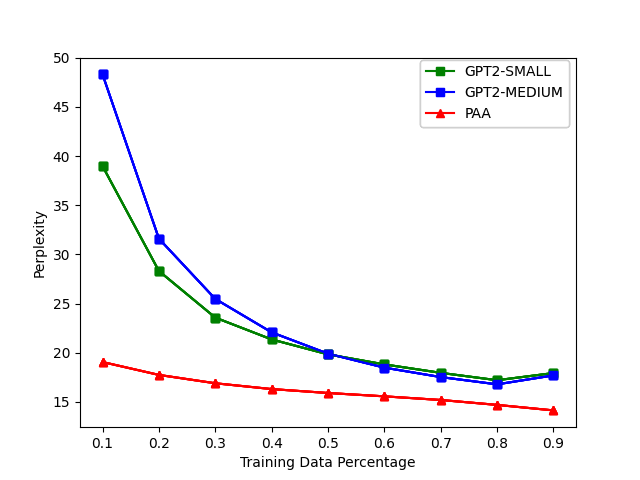}
  \caption{Comparison with GPT2 under low-resource scenario, we sampled 10\% to 90\% of training data to train GPT2-SMALL, GPT2-MEDIUM and PAA.}
  \label{fig:data_efficient}
  \vspace{-1.25em}

\end{figure}

\subsection{Analysis on Data Efficiency}
\label{sec:data_exp}
We further test the data efficiency of PAA compared with the GPT2-SMALL and GPT2-LARGE, a base model for PAA and a larger model than PAA. As Figure \ref{fig:data_efficient} shows, the proposed method can still get 19.05 perplexity when using only 10\% training data, which is significantly better than GPT2's 38.98 and 48.31 under the same circumstance. As the training data is increased to 20\%, PAA achieves 17.74, a lower perplexity than GPT2-SMALL/MEDIUM trained on the full dataset. Then, we increase the ratio to 30\%, and PAA reaches a perplexity of 16.89, which is better than GPT2-LARGE trained on 100\% data. Hence, PAA can achieve a better result in the low-resource regime, especially in the extremely low ratio (e.g., 10\%, 20\%). This demonstrates the effectiveness of PAA's design under this task and shows that the data scarcity problem could be alleviated through PAA.

\begin{table}[t]
\centering
\begin{tabular}{llll}
\hline
Method & PARAMS & PPL $\downarrow$ & F1 $\uparrow$ \\
 \hline
Reddit 2.7B & 2.7B & 18.90 & 12.60 \\
BlenderBot 1 & 2.7B & \textbf{10.20} & 18.30 \\
R2C2 BlenderBot & 2.7B & 10.50 & \textbf{20.50} \\
OPT-175B & 175B & 10.80 & 18.50 \\ 
PAA (Ours) & 254M & 14.03 & 17.36\\
\hline
\end{tabular}
\caption{Automatic evaluation results on the ConvAI2 dataset over large pre-trained language models.}

\vspace{-1em}
\label{tab:baseline3}

\end{table}

\subsection{Comparison with Large Pre-Trained Models}

There are also several large pre-trained language models that have the ability to conduct persona-based dialogue generation. The Reddit 2.7B ~\cite{reddit_dataset} is an unsupervised large-scale language model pre-trained on the Reddit dataset. The BlenderBot 1~\cite{blenderbot} and R2C2 BlenderBot~\cite{blender2} are the supervised conversational language model that are fine-tuned on the ConvAI2 dataset~\cite{convai2}. The OPT-175B ~\cite{OPT} is a recently released large-scale unsupervised language model evaluated on the ConvAI2 dataset.

As shown in Table \ref{tab:baseline3}, with the mega amount of external data and the large-scale parameters within the models, the performance will continue to increase significantly, but the cost of performance gain comes from huge hardware and human resources consumption. Our model is much smaller than the large-scale language model (254M vs. 2.7B) that was fine-tuned over the ConvAI2 dataset while keeping the comparative results to the best results over the large models. Therefore, this result shows that our method is well-considered for this conditioned generation task and offers parametric-efficient performance.
\section{Conclusion}
In this work, we have presented a framework with PAA that balances and regularizes the input sources from persona and context. The adaptive attention dynamically allocates the weights for persona and context information, while the masking removes the redundant information from two input sources. Experimental results on the public dataset demonstrated our framework's superiority in handling the two constraints for response generation. Also, through explicit model design, our approach can still achieve promising performance in the low-resource regime.

There are many potential directions for future research. First, it is fascinating to investigate more fusion and weighting methods. 
Also, filtering the important context from history through a memory module would be an interesting direction, with the potential of reducing computational complexity and enhancing performance by providing key evidence. 

\section*{Acknowledgments}

This work is supported by NSFC key grant 62136005, NSFC general grant 62076118, and Shenzhen fundamental research program JCYJ20210324105000003.

\bibliography{aaai23}

\pagebreak
\section*{Appendix}

\section{Technical Appendix}
\begin{figure*}[ht]
  \includegraphics[width=\textwidth]{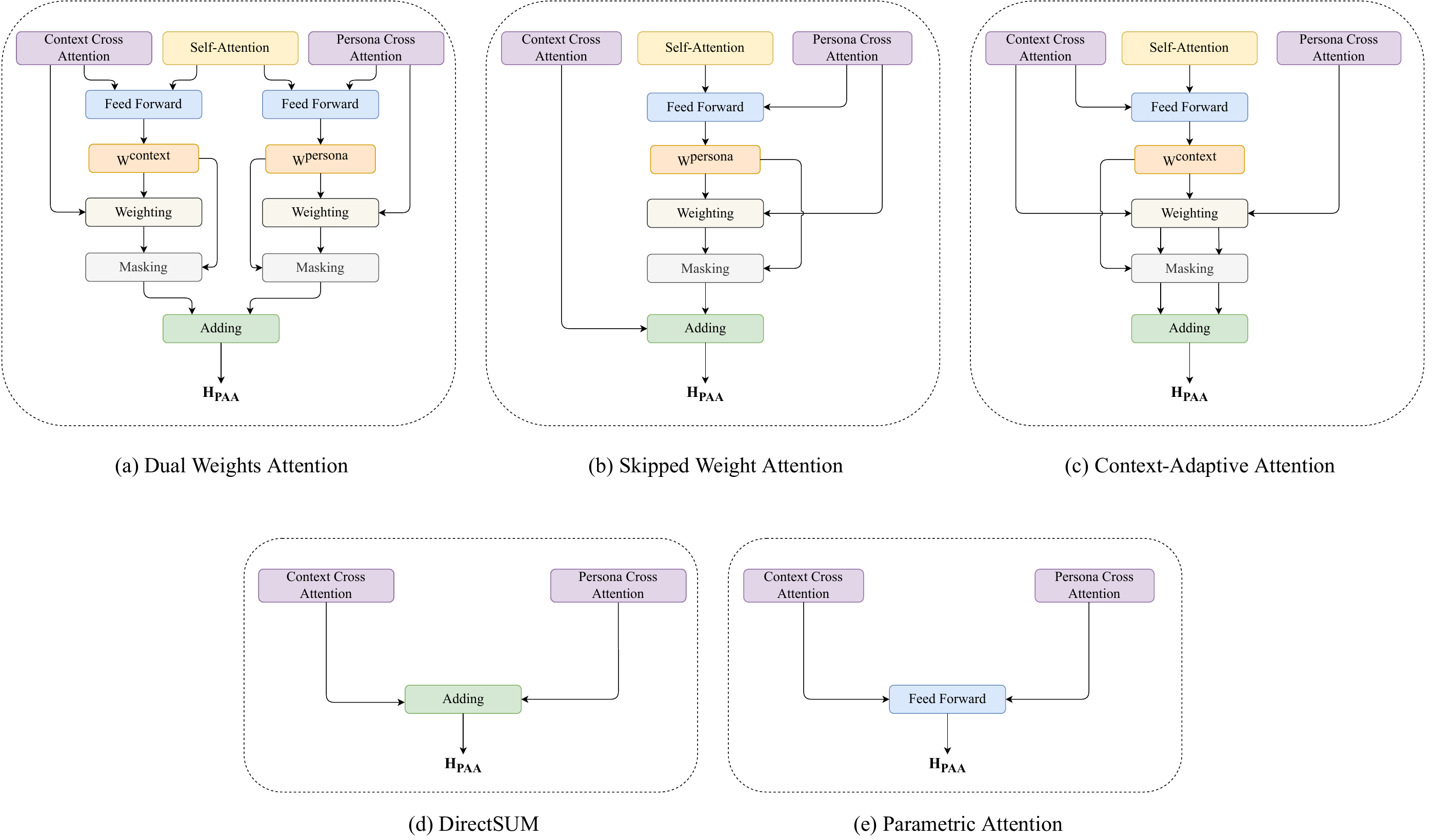}
  \caption{The variants of PAA. (a) Dual Weights Attention: the weights for two cross-attention are calculated separately; (b) Skipped Weight Attention: the weight and mask will only apply to persona cross-attention; (c) Context-Adaptive Attention: the weights and masks inferred from context rather than persona cross-attention; (d) DirectSUM directly sums two cross attention; (E) Parametric Attention processes two cross-attention results via a feed-forward network.}
  \label{fig:variant}
\end{figure*}




\subsection{Variants of Persona-Adaptive Attention}
\label{sec:variant}
In order to explore the effectiveness of the attention mechanism, we also propose five variants of the attention to validate the effectiveness of our proposed method.

\subsubsection{DirectSUM}
In Figure \ref{fig:variant}(d), the two cross-attention results will be directly summed together.

\subsubsection{Parametric Attention (PARAM)}
In Figure \ref{fig:variant}(e), the two cross-attention results will be concatenated as the input to one linear layer, and the output of the linear layer will be regarded as the balanced output.

\subsubsection{Dual Weights Attention (Dual)}
As shown in Figure \ref{fig:variant}(a), we separate the weight calculating linear layer for persona and context cross-attention results correspondingly. This will validate whether the balanced result should come from one weight or two weights. In this design, each linear layer will project its own cross-attention result to the weight and apply the weight to generate the weighted masked results.
\subsubsection{Skipped Weight Attention (Skipped)}
As illustrated in Figure \ref{fig:variant}(b), this variant only apply weighting and masking operation to the persona cross-attention, skipping these operations for the context cross-attention. This will validate the effectiveness of the weight and mask applied to the context-response cross-attention.

\subsubsection{Context-Adaptive Attention (Context)}
In Figure \ref{fig:variant}(c), we calculate the weight from context cross-attention with self-attention instead of persona cross-attention. This variant will validate whether the persona-related information will be more important than the context information in the dialogue generation.

\subsection{Reproduction of Attention-Routing}
\label{sec:re_attn_routing}
Attention-Routing needs a persona weight $\alpha$ to control the strength for persona cross-attention. However, the $\alpha$ comes from a weight predictor that trained from a dataset created by the heuristic script. Since neither the script nor dataset is publicly available, we empirically set the $\alpha$ through grid search that might lead to a sub-optimal result.

In the reproduction of Attention-Routing Design, we add the result of cross-attention as follows
\begin{equation}
\begin{aligned}
h_{AR}=h_R+\alpha o_P + (1-\alpha) O_U + O_U,
\end{aligned}
\end{equation}
where $\alpha$ is a hyper-parameter determined from grid-search. Here, we set the $\alpha=0.2$.

\subsection{Reproduction of Conditioned Transformer Block}
\label{sec:re_cond_trans}
The conditioned transformer block is initially designed for categorical label condition rather than free-form text. Therefore, we modified the structures of the conditioned transformer block to accept the free-form text. In the re-implementation of the conditioned transformer block, we still use two encoders to encode persona and context text:
\begin{equation}
\begin{aligned}
    h_P&=\text{Encoder}_P(I_P),\\
    h_U&=\text{Encoder}_U(I_U),
\end{aligned}
\end{equation}
Then, we apply mean pooling to the encoded persona and context as the conditions represented by the dense vectors
\begin{equation}
\begin{aligned}
    c_P&=\text{Pooling}(h_P),\\
    c_U&=\text{Pooling}(h_U).
\end{aligned}
\end{equation}

We also apply mean/max pooling to the decoder's self-attention result:
\begin{equation}
\begin{aligned}
    c_R&=\text{Pooling}(h_R).
\end{aligned}
\end{equation}

Afterward, inside the conditioned transformer, we do following calculation:
\begin{equation}
\begin{aligned}
    Bias&=\text{Softmax}(\frac{Q_R K_M^\top}{\sqrt{d}})V_M,
\end{aligned}
\end{equation}
where $Q_R$ comes from the linear transformation of $I_R$, and $K_M$ and $V_M$ are from the linear transformations of the concatenation of $[c_P;c_U;c_R]$. The $Bias$ term is added to the self-attention result in the decoder and the remaining are the same as in the conventional transformer block. 
\begin{table}

\begin{tabular}{llll}
\hline
Method & PARAMS & PPL $\downarrow$ & F1$\uparrow$ \\
\hline
Cond-Aware (Mean) & 254M & 120.57 & 0.09 \\
Cond-Aware (Max) & 254M & 88.05 & 0.15 \\
PAA (Ours) & 254M & 14.03 & 17.36\\ \hline
\end{tabular}
\caption{Automatic evaluation results on ConvAI2 dataset over our implemented approach. Cond-Aware (Mean) means the condition-aware transformer with MeanPooling, and Cond-Aware (Max) means the condition-aware transformer with MaxPooling.}

\end{table}

\section{Code and Data Appendix}

\subsection{PAA PyTorch Pseudo Code}
The PyTorch-style Pseudo code is listed in Listing \ref{lst:code}.

\begin{listing}[h]%
\caption{PyTorch-style Pseudo-code for PAA Training}%
\label{lst:code}%
\begin{lstlisting}[language=python]
# x1: tokenized persona
# x2: tokenized context
# t: target response
# get persona encoder representation
e1 = encoder_persona(x1)
# get context encoder representation
e2 = encoder_context(x2)
# get embedding result from decoder
d = emb_decoder([x1, [BOS], t, [EOS]])
# process in decoder block for N_d times
while layer in layers:
    # calculate tau
    tau = len(x2) / (len(x1) + len(x2))
    attn = self_attn(d) + d
    # get persona/context cross-attention results: c1/c2
    c1 = cross_attn(e1, attn)
    c2 = cross_attn(e2, attn)
    # calculate w_1 from c1 & decoder's attn
    w1 = sigmoid(linear([c1; attn]))
    # make w2 complementary to w1
    w2 = 1 - w1
    # obtain mask by w1 & tau
    m1 = where(w1 > tau, 0, 1)
    m2 = where(w1 < 1 - tau, 0, 1)
    # get the H_{PAA} as c3 by weighting and masking
    c3 = w1 * m1 * c1 + w2 * m2 * c2
    attn = c3 + attn
    d = mlp(attn) + attn
pred = softmax(linear(d))[len([x1]):-1]
loss = cross_entropy(pred, t)
loss.backward()

\end{lstlisting}
\end{listing}

\subsection{Performance with Different Seeds}
To test the robustness of our methods, we re-run PAA with the different seed settings. The mean and standard deviation (STD) of performance results are displayed in Table \ref{tab:variance}. We use five seeds to validate our model's performance. 

\begin{table}[h]
\centering
\begin{tabular}{lrr}
\hline
Metric & \multicolumn{1}{l}{Mean} & \multicolumn{1}{l}{STD} \\ \hline
PPL    & 14.0948                  & 0.0422                  \\
F1     & 17.5367                  & 0.0014                  \\
BLEU-1 & 20.5925                  & 0.0081                  \\
BLEU-2 & 4.1264                  & 0.0037                  \\
Dist-1 & 1.3325                   & 0.0004                  \\
Dist-2 & 5.3660                   & 0.0014                  \\ \hline

\end{tabular}
\caption{Performance Mean and Standard Deviation for the PAA experiments on different seeds.}
\label{tab:variance}
\end{table}
\end{document}